\documentclass{article}
\usepackage{spconf,amsmath,xcolor}

\usepackage{enumitem}
\usepackage{amsfonts}
\usepackage{booktabs}
\usepackage{graphicx}
\usepackage{caption}
\usepackage{subcaption}
\usepackage{hyperref}
\setlist{nosep, leftmargin=14pt}

\usepackage{mwe} 
\newcommand{\best}[1]{\textbf{\textcolor{red}{#1}}}
\newcommand{\second}[1]{{\textcolor{blue}{#1}}}

\title{Bridging Classification and Segmentation in Osteosarcoma Assessment via Foundation and Discrete Diffusion Models}
\name{\begin{tabular}{c}
Manh Duong Nguyen$^{1}$, Dac Thai Nguyen$^{1}$, Trung Viet Nguyen$^{2}$, Homi Yamada$^{5}$, \\
Huy Hieu Pham$^{3,4}$, Phi Le Nguyen$^{1, *}$\end{tabular}\thanks{* Corresponding author: \href{mailto:lenp@soict.hust.edu.vn}{lenp@soict.hust.edu.vn}}}

\address{
    $^{1}$Hanoi University of Science and Technology, Hanoi, Vietnam \\
    $^{2}$Pathology Division, Vinmec Times City International Hospital, Vinmec Healthcare System, Hanoi, Vietnam \\
    $^{3}$College of Engineering \& Computer Science, VinUniversity, Hanoi, Vietnam \\
    $^{4}$VinUni-Illinois Smart Health Center, VinUniversity, Hanoi, Vietnam \\
    $^{5}$College of Health Sciences, VinUniversity, Hanoi, Vietnam
}

\begin{document}
%
\maketitle

\begin{abstract}
Osteosarcoma, the most common primary bone cancer, often requires accurate necrosis assessment from whole slide images (WSIs) for effective treatment planning and prognosis. However, manual assessments are subjective and prone to variability. In response, we introduce FDDM, a novel framework bridging the gap between patch classification and region-based segmentation. FDDM operates in two stages: patch-based classification, followed by region-based refinement, enabling cross-patch information intergation. Leveraging a newly curated dataset of osteosarcoma images, FDDM demonstrates superior segmentation performance, achieving up to a 10\% improvement mIOU and a 32.12\% enhancement in necrosis rate estimation over state-of-the-art methods. 
This framework sets a new benchmark in osteosarcoma assessment, highlighting the potential of foundation models and diffusion-based refinements in complex medical imaging tasks.
\end{abstract}

\section{Introduction}
\label{sec:intro}
\vspace{-5pt}
Osteosarcoma is the most common primary bone cancer, with an annual incidence of 4 to 5 cases per million worldwide~\cite{Osteosarcoma2010}. Induction chemotherapy before surgery is standard for osteosarcoma patients~\cite{treatment1997}. The necrosis ratio—defined as the proportion of necrotic tumor tissue in resected samples—correlates strongly with patient outcomes, with a 5-year survival rate exceeding 80\% for patients whose necrosis ratio is over 90\%. However, manual necrosis assessment from H\&E-stained slides is semi-quantitative and prone to observer variability~\cite{manualOsteo1992}, with interclass correlation of only 0.652 among pathologists~\cite{variability2017}.

Deep learning, a subfield of machine learning, has been widely applied for whole slide image (WSI) analysis due to its objectivity and reproducibility~\cite{Srindhi2021, Ho2022}. 
To address technical limitations, researchers often analyze mini-patches cropped from WSIs independently, using two main approaches: patch classification and region segmentation. Patch classification, while effective, often lacks sufficient cross-patch context, leading to suboptimal results. Region segmentation typically requires substantial data to perform well~\cite{Ho2022}. Additionally, most methods rely on traditional CNNs~\cite{Ho2022, Vezakis2023, Walid2023, Aziz2023}, overlooking the potential of foundation models—a limitation particularly impactful for complex cancers like osteosarcoma.

\begin{figure*}[t]
    \centerline{\includegraphics[width=0.88\linewidth]{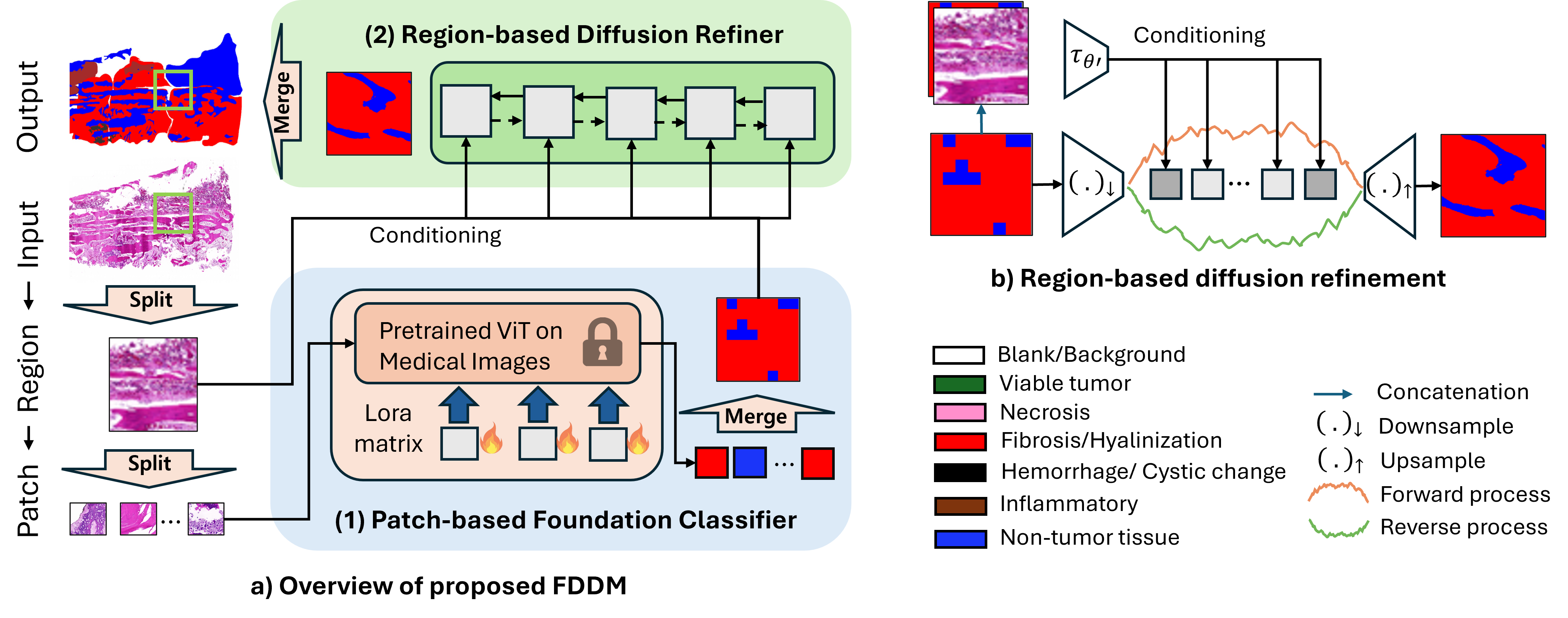}}
    \vspace{-4pt}
    \caption{\textbf{Workflow of FDDM in WSI Segmentation}. Figure 1a overviews our two-stage framework: a patch-based foundation classifier followed by a region-based diffusion refiner for segmentation. Figure 1b details the second stage, where the refiner transforms coarse prediction \( Y_i \) into refined segmentation \( X_i \) through a denoising diffusion process, integrating tissue region \( O = (R_i)_\downarrow \) and \( Y_i \) to guide spatial embedding in the hidden state.}
    \label{fig:fddm_all}
    \vspace{-8pt}
\end{figure*}

To address these challenges, we introduce FDDM, a novel framework that leverages \textbf{F}oundation and \textbf{D}iscrete \textbf{D}iffusion \textbf{M}odel to bridge classification and segmentation, thus enhancing overall performance in WSI analysis. FDDM operates in two stages: an initial patch classification phase, followed by region-based refinement. An overview of FDDM is shown in Figure~\ref{fig:fddm_all}. The main contributions of this study are as follows:
\begin{itemize}
    \item We introduce FDDM, a novel framework for WSI segmentation. FDDM leverages a foundation model for efficient patch classification on large pathology images while preserving global context through region-based refinement. This is the first study to bridge classification and segmentation in WSI analysis using foundation and diffusion models. Source code can be found in  \url{https://github.com/nmduonggg/FDDM}.
    \item We present two comprehensive osteosarcoma datasets: one with nearly 900,000 annotated mini-patches for patch classification and another with approximately 51,000 samples for region-based segmentation, providing substantial resources for advancing cancer assessment.
    \item Our experiments thoroughly evaluate the proposed method against various patch-based classification and region-based segmentation techniques. 
    FDDM outperforms these approaches by up to 10\% in mIOU and 32.12\% in necrosis rate estimation.
\end{itemize}
\vspace{-10pt}

\section{Methodology}
\label{sec:methodology}
\subsection{Overview of FDDM}
In WSI analysis, the two main approaches—patch classification and region segmentation—each have limitations: patch classification lacks cross-patch context, while region segmentation depends on large datasets for optimal performance. To address these challenges, we propose FDDM, a dual-phase framework. The first phase uses patch-based classification to leverage the abundance of small patches, while the second phase applies region-based refinement to enhance coarse segmentations derived from aggregated patches. This design enables the refiner to utilize guidance from classification, eliminating the need for direct inference from raw tissue images, as required in conventional segmentation networks.

Technically, given a whole slide image $\boldsymbol{I}$, FDDM splits it into $N$ large regions $\{R_{i}\}_{i=1}^N$ of a fixed size, i.e., $h \times w$, which further are divided into a set of $K$ smaller patches $\{P_{ij}\}_{j=1}^K$ with shape of $h' \times w'$. Following conventional patch classification settings, a patch $P_{ij}$ is fed into a patch classifier $\mathcal{F}_W$ with parameter $W$, mathematically expressed as $Y_{ij} = \mathcal{F}_W(P_{ij})$, before being merged into an intermediate classification mask $Y_i = \bigcup \{P_j\}_{j=1}^K$. Subsequently, these patch-wise classification masks are smoothened and corrected by utilizing a region-based refiner $\mathcal{R}$ as $X_i = \mathcal{R}(Y_i)$.
Based on generated refined classification masks $\{X_i\}_{i=1}^N$, FDDM merges them to generate the final segmentation mask.
\vspace{-8pt}

\subsection{Patch-based Foundation Classifier} \label{subsec:patch_classification}


Previous studies have primarily used CNNs~\cite{Ho2022, arunachalam2019} to learn \(\mathcal{F}_W\), overlooking the potential of foundation models like Vision Transformer (ViT)~\cite{dosovitskiy2021imageworth16x16words} due to its high computational cost in medical imaging. Hence, we propose to adopt Low-rank Adaptation (LoRA)~\cite{hu2021lora} to finetune ViT, i.e., UNI~\cite{chen2024uni} in particular.

Let \(d\) and \(k\) denote the hidden dimensions of model parameters. For a weight matrix \(W \in \mathbb{R}^{d \times k}\), the updates are represented as a low-rank decomposition: \(W_{t+1} = W_{t} + \Delta W = W + BA\), where \(B \in \mathbb{R}^{d \times r}\) and \(A \in \mathbb{R}^{r \times k}\), with \(r\) as the rank of decomposition. Only \(B\) and \(A\) are updated during training, while the pre-trained \(W\) remains fixed. Thus, LoRA replaces the standard fine-tuning update with a low-rank update:
\begin{equation}
    B \leftarrow B - \eta \nabla_B \mathcal{L}_{\text{cls}}(BA), \quad A \leftarrow A - \eta \nabla_A \mathcal{L}_{\text{cls}}(BA),
\end{equation}
where \(\eta\) is the learning rate, and \(\mathcal{L}_{\text{cls}}\) is the patch classification objective, i.e., standard cross-entropy in this study.


\begin{figure*}[t]
\centerline{\includegraphics[width=0.856\linewidth]{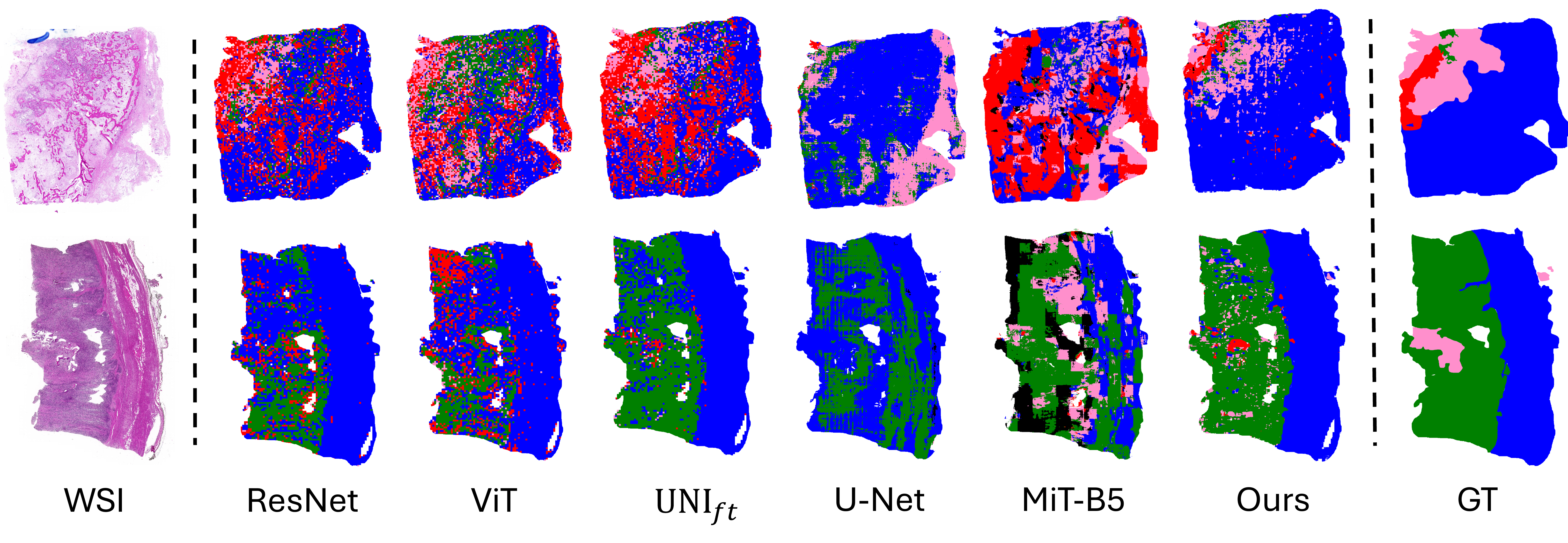}}
    \vspace{-6pt}
    \caption{\textbf{Qualitative results of different methods on testing WSIs.} FDDM’s final predictions are visually smoother and more accurate than other baselines. Notably, UNI$_{ft}$ with the integrated refinement phase, which is indeed FDDM, significantly improves segmentation results over the original. }
    \label{fig:qualitative_results}
    \vspace{-12pt}
\end{figure*}


\vspace{-5pt}
\subsection{Region-based Diffusion Refiner}

In this study, we employ the Brownian Bridge Diffusion Model (BBDM)~\cite{li2023bbdm} to convert patch-wise classification masks, \(\{Y_{i}\}_{i=1}^N\), into segmentation masks, \(\{X_{i}\}_{i=1}^N\). To ensure the discrete nature of segmentation tasks, we apply diffusion directly to probability outputs.

For simplicity, we denote the target segmentation mask and classification mask for region \(R_i\) as \(x := X_i\) and \(y := Y_i\), each shaped \((C, h, w)\), where \(C\) is the class count. During training, FDDM learns to map \(y\) to \(x\) via the conditional Brownian Bridge process, informed by the downsampled tissue region \(O := (R_i)_\downarrow\). 


\noindent \textbf{Forward Process.}
Following \cite{li2023bbdm}, given an initial state \(\boldsymbol{x}_0\) and a destination state \(\boldsymbol{y}\), the intermediate state \(\boldsymbol{x}_t\) at timestep \(t\) can be computed in discrete form as \(\boldsymbol{x}_t = (1 - m_t) \boldsymbol{x}_0 + m_t \boldsymbol{y} + \sqrt{\delta_t} \epsilon_t\). Here, \(m_t = t / T\), where \(T\) is the total number of diffusion steps, \(\delta_t\) is the variance of the Brownian Bridge process, and \(\epsilon\) represents Gaussian noise, i.e., \(\epsilon \sim \mathcal{N}(\mathbf{0}, \mathbf{I})\).
Thus, the forward process can be defined as 
\(q_{BB}(\boldsymbol{x}_t | \boldsymbol{x}_0, \boldsymbol{y}) = \mathcal{N}(\boldsymbol{x}_t; (1 - m_t) \boldsymbol{x}_0 + m_t \boldsymbol{y}, \delta_t \mathbf{I})\).
To determine the transition probability between consecutive steps during training, we use the formula provided in \cite{li2023bbdm}.
\\
\noindent \textbf{Reverse Process.}
The reverse process begins with \( \boldsymbol{x}_T = \boldsymbol{y} \). Embedding spatial characteristics and enforcing supervision from prior classifications, we condition the refinement by integrating original tissue region \( \boldsymbol{O} \) and classification mask \( \boldsymbol{y} \) into the hidden state \( \boldsymbol{x}_t \) at each timestep \( t \). Information from \( \boldsymbol{O} \) and \( \boldsymbol{y} \) is concatenated along the channel dimension, processed through a convolutional network \( \boldsymbol{\tau}_{\theta'} \), and channel-wise concatenated with \( \boldsymbol{x}_t \) as follows:
\vspace{-2pt}
{\small
\begin{equation*}
\begin{aligned}
    &p_\theta(\boldsymbol{x}_{t-1} | \boldsymbol{x}_t, \boldsymbol{y}, \boldsymbol{O}) = \mathcal{N} (\boldsymbol{x}_{t-1}; \mu_\theta (\boldsymbol{x}_t, \boldsymbol{y}, \boldsymbol{O}, t ),
    \tilde{\delta}_t \mathbf{I}), \nonumber \\
   &\mu_\theta (\boldsymbol{x}_t, \boldsymbol{y}, \boldsymbol{O}, t) = c_{x t} \boldsymbol{x}_t + c_{y t} \boldsymbol{y}
   + c_{\epsilon t} \boldsymbol{\epsilon}_\theta (\boldsymbol{\tau}_{\theta'}(\boldsymbol{y} \oplus \boldsymbol{O}), \boldsymbol{x}_t, t), 
\label{eq:reverse_process}
\end{aligned}
\vspace{-8pt}
\end{equation*}}\\
where $\mu_\theta (\boldsymbol{x}_t, \boldsymbol{y}, \boldsymbol{O}, t)$ represents the predicted mean and $\tilde{\delta}_t$ represents the variance of the distribution at time step $t$, respectively; $c_{x t}$, $c_{y t}$ and $c_{\epsilon t}$ are non-trainable factors 
as described in \cite{li2023bbdm}.
In this work, we extend the foundational principles of Denoising Diffusion Implicit Models (DDIM)~\cite{song2020denoising}, to enhance the sampling process efficiency, similar to \cite{li2023bbdm}.

\vspace{-5pt}
\subsection{Refinement Training Objective}

\noindent \textbf{Transition Loss.} $\mathcal{L}_{trans}$ optimizes the disparity between the predicted and observed transition distribution of the intermediate states in the forward diffusion process, as indicated in \cite{li2023bbdm}. This is achieved by learning the noise distribution $\epsilon_\theta$ 
parameterized by $\theta$ via minimizing the simplified Evidence Lower Bound (ELBO) as follows:
\begin{align}
\label{eq:transition_loss}
    \vspace{-8pt}
    \mathcal{L}_{trans}  = &\mathbb{E}_{\boldsymbol{x}_0, \boldsymbol{y}, \boldsymbol{\epsilon}} [c_{\epsilon t}||m_t(y-x_o) \nonumber \\ 
    &+ \sqrt{\sigma_t}\boldsymbol{\epsilon} - \boldsymbol{\epsilon}_\theta (\boldsymbol{\tau}_{\theta'}(\boldsymbol{y} \oplus \boldsymbol{O}), x_t, t)||^2].
    \vspace{-8pt}
\end{align}
Equation~\ref{eq:transition_loss} highlights the inclusion of information from both tissue and classification masks, intuitively.

\noindent \textbf{Segmentation Loss.} $\mathcal{L}_{seg}$
is proposed to improve the accuracy of the final generated mask, addressing the diffusion model's typical focus on minimizing transition prediction error rather than aligning with ground truth. \cite{li2023bbdm} shows that the Markovian nature of BBDM’s forward process allows for an approximate estimation of the final prediction at any time step, as defined in Equation~\ref{eq:x0_recon}.
\vspace{-4pt}
\begin{equation} \label{eq:x0_recon}
    \boldsymbol{\hat{x}}_0 = \frac{\boldsymbol{x}_t - m_t \boldsymbol{y} + \sqrt{\sigma_t} \, \boldsymbol{\epsilon}_\theta (\boldsymbol{\tau}_{\theta'}(\boldsymbol{y} \oplus \boldsymbol{O}), x_t, t)}{1 - m_t}.
    \vspace{-4pt}
\end{equation}
This formulation enables minimizing the disparity between the refiner’s final output and the ground-truth segmentation mask through a conventional cross-entropy loss:
\vspace{-4pt}
\begin{equation}
    \mathcal{L}_{seg} = - \boldsymbol{x}_0 \log(\boldsymbol{\hat{x}}_0).
    \vspace{-4pt}
\end{equation}
Intuitively, $\mathcal{L}_{seg}$ as a guiding mechanism, ensuring that each denoising step in the reverse process should contribute to producing an accurate segmentation mask, rather than solely focusing on denoising as seen in continuous diffusion frameworks~\cite{li2023bbdm, song2020denoising}. Together, these two loss functions form the final objective for training the region-based refiner, resulting in the refinement objective:
\vspace{-4pt}
\begin{equation}
    \mathcal{L}_{ref} = \mathcal{L}_{trans} + \lambda \mathcal{L}_{seg},
    \vspace{-4pt}
\end{equation}
where \(\lambda\) is a hyperparameter that controls the impact of $\mathcal{L}_{seg}$ on the final loss and is set to 1.0 by default.

\vspace{-10pt}


\section{Evaluation}
\label{sec:evaluation}
\subsection{Dataset and Experimental Settings}

\subsubsection{Data Preparation}
To address the scarcity of osteosarcoma-focused datasets, we collected and processed 160 WSIs from various bone regions of patients at Vinmec International Healthcare System in 2024. Bone samples were de-calcified, H\&E stained, digitized, and annotated by two pathologists to categorize regions into seven classes: Viable Background (BG), Viable Tumor (VT), Necrosis (NC), Fibrosis/Hyalination (FH), Hemorrhage/Cystic Change (HC), Inflammatory (IF), and Non-tumor Tissue (NT). We derived two datasets for different experimental needs:

\noindent \textbf{Patch-based Dataset.} Each WSI was divided into 256 × 256 pixel tiles, resulting in 892,073 tissue tiles after filtering. Each tile was labeled by its dominant class, with class distribution shown in Table~\ref{tab:dts_distribution}.

\noindent \textbf{Region-based Dataset.} Using similar preprocessing, tiles were cropped at \(256 \times 2^k\) pixel sizes to capture broader patch relationships but reducing instances. This dataset includes 51,264 tiles and is tailored for segmentation. For this study, \(k\) was set to 3 to balance cross-patch information and computational cost.
\begin{table}[t]
\centering
\caption{Class distribution in our proposed patch-based datasets.}
\vspace{-5pt}
\label{tab:dts_distribution}
\resizebox{\linewidth}{!}{%
\begin{tabular}{@{}l|ccccccc@{}}
\toprule
\textbf{Dataset} & \textbf{BG} & \textbf{VT} & \textbf{NC} & \textbf{FH} & \textbf{HC} & \textbf{IF} & \textbf{NT} \\ \midrule
\textbf{Train}            & 20,407      & 127,744     & 105,900     & 142,522     & 8,037       & 2,276       & 306,770     \\
\textbf{Valid }           & 2,550       & 15,968      & 13,237      & 17,815      & 1,004       & 284         & 38,346      \\
\textbf{Test}             & 2,552       & 15,969      & 13,238      & 17,816      & 1,006       & 285         & 38,347      \\ \midrule
\textbf{Total} &
  \begin{tabular}[c]{@{}c@{}}25,509\\ (2.86\%)\end{tabular} &
  \begin{tabular}[c]{@{}c@{}}159,681\\ (17.90\%)\end{tabular} &
  \begin{tabular}[c]{@{}c@{}}132,327\\ (14.84\%)\end{tabular} &
  \begin{tabular}[c]{@{}c@{}}178,153\\ (19.97\%)\end{tabular} &
  \begin{tabular}[c]{@{}c@{}}10,047\\ (1.13\%)\end{tabular} &
  \begin{tabular}[c]{@{}c@{}}2,845\\ (0.32\%)\end{tabular} &
  \begin{tabular}[c]{@{}c@{}}383,463\\ (42.99\%)\end{tabular} \\ \bottomrule
\end{tabular}%
}
\vspace{-6pt}
\end{table}
\vspace{-8pt}
\subsubsection{Experimental Settings}

\noindent \textbf{Baseline Methods}.
We evaluate the proposed FDDM against baseline methods from two common approaches in WSI analysis: patch classification and region segmentation. For patch classification, we employ ResNet-101~\cite{he2015deepresiduallearningimage}, ViT~\cite{dosovitskiy2021imageworth16x16words}, and our foundation classifier (UNI$_{ft}$). In the realm of region segmentation, we utilize state-of-the-art methods (U-Net~\cite{ronneberger2015u} and MiT-B5~\cite{xie2021segformersimpleefficientdesign}). The classification networks are trained on a patch-based dataset, while the segmentation methods are applied to a region-based dataset.

\noindent \textbf{Evaluation Metric}. 
We evaluate all methods based on two criteria: segmentation performance (mIOU, Precision, Recall) and pathological accuracy, following \cite{Ho2022}, by comparing the total necrosis rate (TNR) from pathology reports (\( r_{PR} \)) with our model's estimate (\( r_{DL} \)). The necrosis rate is calculated as follows:
\begin{equation*}
r_{DL} = \frac{p_{NC} + p_{FH} + p_{HC} + p_{IF}}{p_{VT} + p_{NC} + p_{FH} + p_{HC} + p_{IF}},    
\vspace{-1pt}
\end{equation*}
where \( p_{c} \) denotes the pixel count in WSIs for class \( c \). We hypothesize that the model’s necrosis ratio will align closely with expert assessments, reporting the absolute difference \( |r_{PR} - r_{DL}| \) to demonstrate effectiveness.
\vspace{-6pt}
\subsection{Algorithmic Segmentation Performance}
\begin{table}[t]
\small
\centering
\caption{Quantitative results of different methods in common segmentation metrics. The best and second-based results are highlighted by \textcolor{red}{\textbf{red}} and \textcolor{blue}{blue}, respectively. The higher is better. }
\vspace{-5pt}
\label{tab:segment_results}
\resizebox{\linewidth}{!}{%
\begin{tabular}{@{}l|lll|ll||l@{}}
\toprule
\textbf{Metric}    & ResNet~\cite{he2015deepresiduallearningimage}  & ViT~\cite{dosovitskiy2021imageworth16x16words}    & UNI$_{ft}^{*}$     & U-Net~\cite{ronneberger2015u} & MiT-B5~\cite{xie2021segformersimpleefficientdesign} & \textbf{FDDM (Ours)}        \\ \midrule
\textbf{mIOU }      & 33.12\% & 32.03\% & \second{40.18\%} &    34.22\%    &  31.31\%        & \best{44.91\%} \\
\textbf{Precision} & 45.52\% & 47.13\% & \second{51.56\%} &    49.69\%   &    42.78\%       & \best{54.55\%} \\
\textbf{Recall}    & 69.67\% & 63.45\% & \second{71.45\%} &   62.99\%    &    65.26\%       & \best{72.37\%} \\ \bottomrule
\end{tabular}%
}
\begin{minipage}{\linewidth}
\centering
\footnotesize
\vspace{3pt}
$^*$UNI$_{ft}$ is a version fine-tuned with LoRA of UNI~\cite{chen2024uni}.
\end{minipage}
\vspace{-24pt}
\end{table}
\noindent \textbf{Qualitative Results.}
Figures~\ref{fig:qualitative_results} compares the multiclass segmentation predictions on WSIs of different methods, highlighting that patch-based classification outperforms region-based segmentation due to its larger dataset volume from smaller crop sizes. Additionally, FDDM’s predictions on testing WSIs exhibit enhanced smoothness and accuracy since it integrates both patch-based and region-based datasets.

\noindent \textbf{Quantitative Results.}
Table~\ref{tab:segment_results} indicates that FDDM outperforms other baselines in all metrics, with recall of 72.37\% surpassing all others by up roughly 10\%. Interestingly, without the region-based refinement, our foundation classifier UNI$_{ft}$ consistently outperforms others by up to 8.87\%, 8.78\%, 8.46\% in mIOU, Precision, Recall respectively, highlighting the effectiveness of foundation model in WSI analysis.
\begin{table}[t]
\centering
\caption{\textbf{Absolute difference of different methods in necrosis rate estimation}. The best and second-best results are highlighted by \textcolor{red}{\textbf{red}} and \textcolor{blue}{blue}, respectively. The lower is better.}
\vspace{-5pt}
\label{tab:rate_results}
\resizebox{\linewidth}{!}{%
\begin{tabular}{@{}l|ccccccc@{}}
\toprule
\textbf{Method} &
  \textbf{VT} &
  \textbf{NC} &
  \textbf{FH} &
  \textbf{HC} &
  \textbf{IF} &
  \textbf{NT} &
  \textbf{TNR} \\ \midrule
ResNet~\cite{he2015deepresiduallearningimage} & 7.32 \% & 15.95\% & 10.04\% & \second{0.10\%} & 0.48\% & 9.21\%  & 29.15\% \\
ViT~\cite{dosovitskiy2021imageworth16x16words}   & 11.58\% & 9.45\%  & 20.75\% & 0.43\% & 0.03\% & 14.13\% & 29.92\% \\
UNI$_{ft}^{*}$    & \second{3.64\%}  & \second{6.99}\% & 12.35\% & \second{0.10\%} & \second{0.02\%} & \second{8.42\%} & \second{10.65\%} \\ \midrule
U-Net~\cite{ronneberger2015u}    & 19.28\%  & 21.29\% & \second{5.33\%} & 0.63\% & \best{0.00\%} & 33.14\% & 35.96\% \\
MiT-B5~\cite{xie2021segformersimpleefficientdesign} &
  {12.57\%} &
  {14.88\%} &
  {13.81\%} &
  {9.99\%} &
  {\second{0.02\%}} &
  {15.30\%} &
  {34.27\%} \\ \midrule \midrule
\textbf{FDDM (Ours)} &
  \best{2.29\%} &
  \best{5.79\%} &
  \best{8.09\%} &
  \best{0.07\%} &
  \best{0.00\%} &
  \best{6.07\%} &
  \best{4.17\%} \\ \bottomrule
\end{tabular}%
}
\vspace{6pt}
\begin{minipage}{\linewidth}
\centering
\footnotesize
\vspace{3pt}
$^*$UNI$_{ft}$ is a version fine-tuned with LoRA of UNI~\cite{chen2024uni}.
\end{minipage}
\vspace{-32pt}
\end{table}

\vspace{-10pt}
\subsection{Necrosis Ratio Assessment}
Table~\ref{tab:rate_results} indicates that FDDM consistently outperforms other methods across all categories in general. The proposed approach achieves the best performances in VT (2.29\%), NC (5.79\%), FH (8.09\%), HC (0.07\%), and NT (6.07\%), showcasing its superior accuracy in estimating necrosis rates. The most significant differences are seen in TNR, where FDDM's results are notably better than those of remaining baselines by up to 33\%. This highlights FDDM's ability to provide more precise and reliable estimations, making it the most efficient solution among the compared methods in supporting pathologists in real-world scenarios.


\vspace{-6pt}

\section{Discussion and Conclusion}
\label{sec:conclusion}

This study proposes FDDM, a two-stage framework for osteosarcoma assessment via WSI analysis, combining model-driven patch classification with a diffusion refiner to optimize segmentation. Incorporating cross-patch context, FDDM overcomes limitations in isolated methods, reducing dependency on extensive annotations and clinician input. Experiments on two osteosarcoma datasets, with 900,000 patches for classification and 51,000 for segmentation, establish a rigorous benchmark for evaluating existing methods.


Results demonstrate that FDDM outperforms current methods in segmentation metrics and necrosis rate accuracy, highlighting its potential to improve diagnostic precision in histopathology. In conclusion, FDDM offers a scalable, efficient solution for automated cancer assessment, providing a base for computational pathology. Future work may focus on reducing computational demands and broadening FDDM’s application to other cancer types, amplifying its role in automated histopathological analysis across clinical settings.



\vfill
\pagebreak

\bibliographystyle{IEEEbib}
\bibliography{strings,refs}

\end{document}